\documentclass[10pt,twocolumn,letterpaper]{article}

\usepackage{iccv}              % To produce the CAMERA-READY version
\usepackage{multirow}
\usepackage{appendix}
\usepackage{float}
\usepackage[accsupp]{axessibility}
\usepackage{marvosym}

\definecolor{iccvblue}{rgb}{0.21,0.49,0.74}
\usepackage[pagebackref,breaklinks,colorlinks,allcolors=iccvblue]{hyperref}

\def\paperID{2058} % *** Enter the Paper ID here
\def\confName{ICCV}
\def\confYear{2025}

\newcommand{\tablesize}{\footnotesize}

\definecolor{ceiling}{RGB}{214,  38, 40}   %
\definecolor{floor}{RGB}{43, 160, 4}     %
\definecolor{wall}{RGB}{158, 216, 229}  %
\definecolor{window}{RGB}{114, 158, 206}  %
\definecolor{chair}{RGB}{204, 204, 91}   %
\definecolor{bed}{RGB}{255, 186, 119}  %
\definecolor{sofa}{RGB}{147, 102, 188}  %
\definecolor{table}{RGB}{30, 119, 181}   %
\definecolor{tvs}{RGB}{160, 188, 33}   %
\definecolor{furniture}{RGB}{255, 127, 12}  %
\definecolor{objects}{RGB}{196, 175, 214} %

\title{EmbodiedOcc: Embodied 3D Occupancy Prediction \\ for Vision-based Online Scene Understanding}

\author{
Yuqi Wu\quad Wenzhao Zheng{\textsuperscript{\Letter}}\quad Sicheng Zuo\quad Yuanhui Huang\quad Jie Zhou\quad Jiwen Lu \\
Department of Automation, Tsinghua University, China \\
{\tt\small wuyq24@mails.tsinghua.edu.cn; wenzhao.zheng@outlook.com} \\
}

\begin{document}

 \twocolumn[{
 \renewcommand\twocolumn[1][]{#1}
 \vspace{-12mm}
 \maketitle
 \vspace{-10mm}
 \begin{center}
    \centering
    \includegraphics[width=\linewidth]{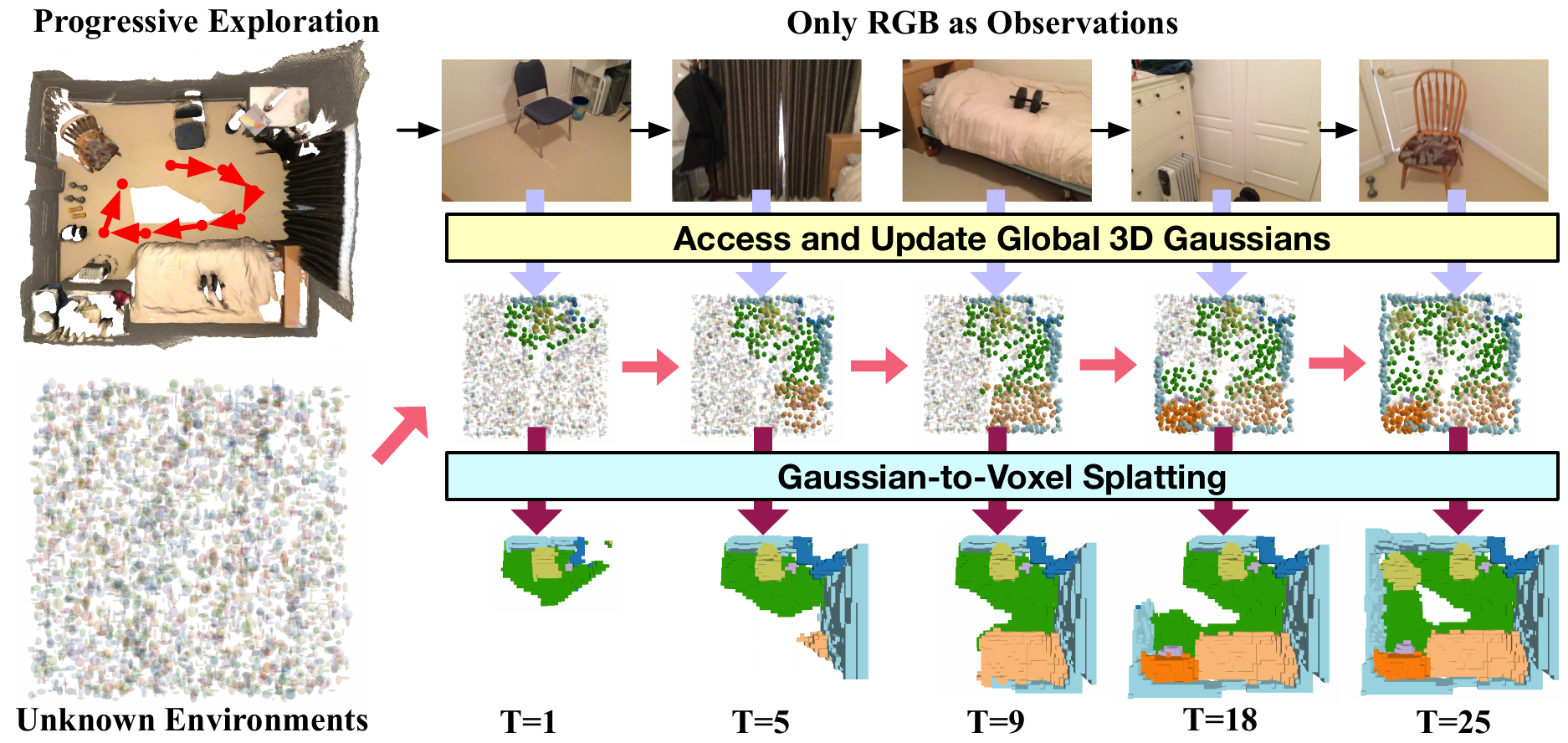}
    \vspace{-7mm}
    \captionof{figure}{
Given streaming monocular RGB inputs, our EmbodiedOcc conducts embodied occupancy prediction in an online manner for indoor scenes.
Different from existing methods which focus on offline perception from monocular images, we focus on the scene-level occupancy prediction from embodied observations.
 We initialize the scene to be explored with uniform 3D semantic Gaussians and progressively update them based on new observations, similar to how humans explore unknown scenes.
 }
 \label{teaser}
 \end{center}%
 }]

\renewcommand{\thefootnote}{\fnsymbol{footnote}}
\footnotetext{\textsuperscript{\Letter}~Corresponding author.}
\renewcommand{\thefootnote}{\arabic{footnote}}

\begin{abstract}
3D occupancy prediction provides a comprehensive description of the surrounding scenes and has become an essential task for 3D perception.
Most existing methods focus on offline perception from one or a few views and cannot be applied to embodied agents that demand to gradually perceive the scene through progressive embodied exploration.
In this paper, we formulate an embodied 3D occupancy prediction task to target this practical scenario and propose a Gaussian-based EmbodiedOcc framework to accomplish it.
We initialize the global scene with uniform 3D semantic Gaussians and progressively update local regions observed by the embodied agent.
For each update, we extract semantic and structural features from the observed image and efficiently incorporate them via deformable cross-attention to refine the regional Gaussians.
Finally, we employ Gaussian-to-voxel splatting to obtain the global 3D occupancy from the updated 3D Gaussians.
Our EmbodiedOcc assumes an unknown (i.e., uniformly distributed) environment and maintains an explicit global memory of it with 3D Gaussians.
It gradually gains knowledge through the local refinement of regional Gaussians, which is consistent with how humans understand new scenes through embodied exploration.
We reorganize an EmbodiedOcc-ScanNet benchmark based on local annotations to facilitate the evaluation of the embodied 3D occupancy prediction task.
Our EmbodiedOcc outperforms existing methods by a large margin and accomplishes the embodied occupancy prediction with high accuracy and efficiency.
Code: \url{https://github.com/YkiWu/EmbodiedOcc}.
\end{abstract}

\vspace{-5mm}    
\section{Introduction}
\label{sec:intro}

With the rapid development of embodied intelligence and active agents~\cite{irshad2022semantically, raychaudhuri2024mopa, lei2024instance}, 3D scene perception~\cite{qi2017pointnet, vu2022softgroup, wang2022cagroup3d, rukhovich2022fcaf3d} has become a crucial task in computer vision. 
Intelligent agents first perceive their surrounding environments and then make decisions based on the perception results. 
Due to the low costs of camera sensors, vision-based 3D occupancy prediction is gaining increasing popularity and produces a comprehensive understanding of both semantics and structures of the scene~\cite{monoscene, surroundocc, tpvformer, huang2025gaussianformer, yu2024monocular}.

%3D perception capabilities required by these agents are diverse, among which 3D occupancy prediction~\cite{monoscene, surroundocc, tpvformer, huang2025gaussianformer, yu2024monocular} is gaining increasing popularity due to its efficiency, uniformity, and scalability.

While vision-based 3D occupancy prediction has made significant progress in outdoor driving scenes~\cite{tpvformer, li2023fb, huang2025gaussianformer, surroundocc, tong2023scene, openoccupancy, pointocc, occworld, occsora, yang2024driving}, the application to indoor scenarios is still challenging due to the diversity and complexity of indoor scenes. 
Most existing methods~\cite{monoscene, Yao_2023_ICCV, yu2024monocular} still focus on local 3D occupancy prediction by integrating semantic and depth information extracted from the visual inputs. 
However, different from outdoor scenarios, it is important to obtain a global understanding of the room for indoor scenarios, as it usually requires multiple traversals for embodied agents.
%these works are inconsistent with the core requirements of embodied agents. 
Also, it is more practical to progressively explore and update the global occupancy of the 3D scene in an online manner from embodied vision-based observations with different positions and perspectives.

%A more promising active agent should be capable of progressively exploring and updating the global occupancy of a 3D scene with the change of its position and perspective, just like most humans can effortlessly accomplish.

To bridge this gap, we formulate a new embodied 3D occupancy prediction task to evaluate the ability to progressively explore an unknown scene using only visual inputs.
We propose an EmbodiedOcc framework based on Gaussian memories to accomplish this task, considering the explicity and structural nature of 3D Gaussians.
We initialize the global scene with uniform 3D semantic Gaussians and progressively update the Gaussians within the field of view observed by the agent. 
Throughout the exploration process, we maintain an explicit global memory of 3D Gaussians as the global understanding and derive the global 3D occupancy with Gaussian-to-voxel splatting~\cite{huang2025gaussianformer}. 
Specifically, we propose a structure-aware local refinement module to update the relevant Gaussians within the current frustum.
%for each update, extracting semantic features from the current input and using these features to update the Gaussians within the current frustum.
% incorporating them via deformable cross-attention. 
% These well-integrated features are used to update the Gaussians within the current frustum. 
We employs a simple yet effective depth-aware branch to introduce explicit structural information for each Gaussian, ensuring the update of these Gaussians to better align with the global representation.
%Based on this effective local prediction module, we update the Gaussian memory progressively.
During the continuous exploration, we read out Gaussians within the current frustum from the memory as inputs to the local module for refinement.
We assign high confidence values for updated Gaussians and use them to reweight information from the memory and the current input.
% and those updated before can provide information of high confidence for this update. 
%We thus generate a confidence value for each Gaussian and use this value to integrate information from both the memory and the current input.
This ensures the consistency of the 3D representation during the fusion and update process.
%, which actually benefits from the physical meaning and structural information of Gaussians.
We reorganize an EmbodiedOcc-ScanNet benchmark for the embodied 3D occupancy prediction task based on the locally annotated Occ-ScanNet dataset~\cite{dai2017scannet, wu2020scfusion, yu2024monocular}.
% to train and evaluate our proposed framework. 
Experiments show that our EmbodiedOcc outperforms existing methods by a large margin and accomplishes embodied occupancy prediction with high accuracy and efficiency.

\section{Related Work}
\label{sec:formatting}

\textbf{3D Occupancy Prediction.}
Benefiting from its compactness and versatility, 3D occupancy prediction based on multi-view images or additional 3D information~\cite{tpvformer, song2017semantic, surroundocc, li2023voxformer, Cai_2021_CVPR, selfocc, huang2025gaussianformer} has gained great popularity over the last few years. 
MonoScene~\cite{monoscene} was the first to derive 3D occupancy prediction from a single image, propelling the original 3D Semantic Scene Completion (SSC)~\cite{song2017semantic, dai2018scancomplete, li2019depth, li2019rgbd, garbade2019two, rist2021semantic} into a more challenging stage with vision-only inputs and more universal scenarios (both indoor and outdoor scenes).
Subsequent works~\cite{Yao_2023_ICCV, yu2024monocular} further focused on addressing the depth ambiguity in this monocular setting. 
However, most of these efforts were confined to local and offline prediction. 
SCFusion~\cite{wu2020scfusion} proposed an incremental framework based on RGB-D inputs. 
EmbodiedScan~\cite{embodiedscan, lyu2024mmscan} introduced an offline global prediction framework using multi-modal sequential inputs. 
Differently, the proposed embodied 3D occupancy prediction aims at online prediction from RGB-only inputs, which is more challenging and practical.

\textbf{Online 3D Scene Perception.}
Accurate comprehension of 3D scenes is an indispensable capability for embodied agents, such as 3D occupancy prediction~\cite{monoscene, yu2024monocular} and object detection~\cite{qi2019deep, wang2022cagroup3d, kolodiazhnyi2024unidet3d}.
Most existing works on indoor 3D scene perception~\cite{qi2017pointnet, graham20183d, yi2019gspn, vu2022softgroup} take pre-acquired and reconstructed 3D data as inputs and perceive the scene in an offline manner.
To achieve online perception, Online3D~\cite{xu2024online} introduced an adapter-based model that equips mainstream offline frameworks with the competence to perform online scene perception, enabling the process of real-time RGB-D sequences.
However, this framework still requires depth information as inputs and mainly targets point segmentation and 3D detection. 
Differently, we target online vision-based 3D occupancy prediction which can provide a more comprehensive understanding of the scene.
%In a more general embodied scenario, real-time monocular visual input for scene perception can further advance the research on embodied agents.

\begin{figure*}[!t]
\centering
\includegraphics[width=\textwidth]{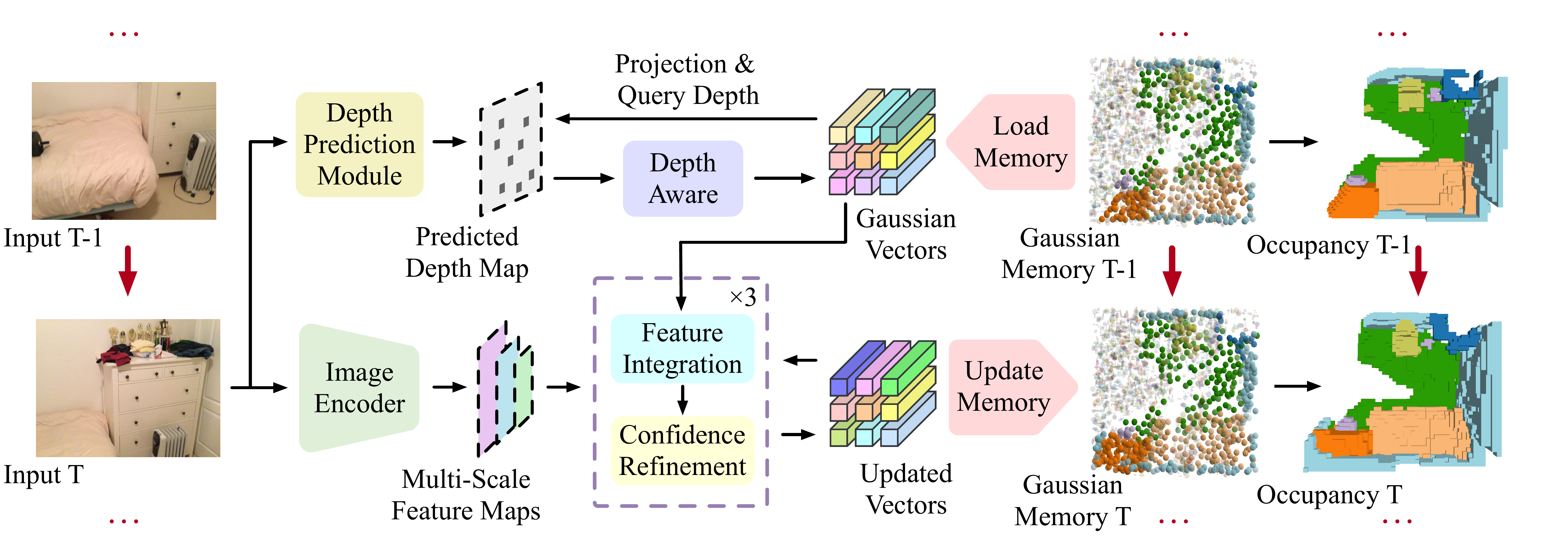}
\vspace{-7mm}
\caption{\textbf{Framework of our EmbodiedOcc for embodied 3D occupancy prediction.}
We maintain an explicit global memory of 3D Gaussians during the exploration of the current scene.
For each update, the Gaussians within the current frustum are taken from the memory and updated using
semantic and structural features extracted from the monocular RGB input.
Each Gaussian has a confidence value to
integrate information from both the memory and the current input.
Then we detach and put these updated Gaussians back into the memory.
We can obtain the current 3D occupancy prediction using a Gaussian-to-voxel splatting module whenever we need.
}
\label{fig:framework}
\vspace{-5mm}
\end{figure*}

\textbf{3D Gaussian Splatting.}
3D Gaussian Splatting~\cite{kerbl3Dgaussians} uses 3D Gaussians to model a 3D scene and benefits from fast speed and high quality in the field of neural rendering.
The physical characteristics of 3D Gaussians and the splat-based rasterization also motivated rapid advancements in research fields such as scene editing~\cite{guedon2024gaussian, palandra2024gsedit, silva2024contrastive, liu2024infusion}, dynamic scenarios~\cite{sun20243dgstream, Lu_2024_CVPR, gao2024gaussianflow, xiao2024bridging, yang20244d}, and SLAM~\cite{yugay2023gaussian, yan2024gs, deng2024compact, li2025sgs}. 
GaussianFormer~\cite{huang2025gaussianformer} pioneers the application of 3D Gaussians in outdoor 3D occupancy prediction and uses features from multi-view images to update 3D Gaussians, which can be converted into 3D occupancy prediction through a Gaussian-to-voxel splatting module. 
% These Gaussians are ultimately converted into 3D occupancy prediction through an elaborately designed module. 
%Compared to conventional voxel-based methods, using 3D Gaussian representation constitutes a more flexible and efficient approach. 
However, it is still unclear how to employ 3D Gaussians for online global indoor scene understanding from local observations.
We achieve this by designing a Gaussian memory mechanism and progressively updating it with structure-aware interaction.
%In this paper, we will leverage this significant attribute to accomplish embodied occupancy prediction in indoor scenarios.

\section{Proposed Approach}

\subsection{Embodied 3D Occupancy Prediction}

Conventional methods in indoor scenarios for occupancy prediction accepted RGB-D as inputs to predict the semantic occupancy of a 3D scene which requires depth sensors.
%This setting provides the model with ample information for inference. 
%However, it undoubtedly diminishes the comprehension capability of the model in practice.
However, we humans are capable of effortlessly processing the visual information from a single view to obtain 3D perception of our surroundings. 
Recent methods begin to consider endowing models with the same competence, which accept a monocular RGB image as input and derive a 3D occupancy prediction within the current frustum:
\vspace{-2mm}
\begin{equation}
\label{eq:eq1} 
% Y = \mathcal{F}\textsubscript{mono}(I)，
\mathbf{Y}_{mono} = \mathcal{F}_{mono}(\mathbf{\mathit{I_{mono} } } ), 
\end{equation}
\vspace{-4mm}

\noindent
where \(\mathcal{F}_{mono}\) is the proposed monocular prediction model, \({\mathbf{\mathit{I_{mono} } }  \in \mathbb{R}^{H \times W \times 3}}\) and \( {\mathbf{Y}_{mono}  \in \mathbb{R}^{X \times Y \times Z \times C}}\) refer to the monocular RGB input and the obtained 3D occupancy prediction. \({X}\), \({Y}\), \({Z}\) represent the dimensions of the local 3D scene and \({C}\) represents the total number of semantics.

This is only the initial step towards practical scenarios. 
The essence of human intelligence is the capacity to analyze and respond immediately based on real-time perception of the surroundings. 
Correspondingly, superior embodied agents are anticipated to process egocentrically gathered real-time visual input to update the 3D occupancy prediction of the current scene. 
% Only when equipped with this capability can the execution of downstream tasks based on real-time perception becomes viable.
This capability facilitates the execution of downstream tasks based on real-time perception.

Motivated by this, we propose an embodied 3D occupancy prediction task in this paper.
Let \(
\mathcal{X}_{t} = \{\mathit{x}_{1}, \mathit{x}_{2}, ..., \mathit{x}_{t} \}
\) be an RGB sequence and the corresponding extrinsics collected by the embodied agent up to the present, where \(
\mathit{x}_{t} = (\mathit{I}_{t}, \mathit{M}_{t}) , \mathit{I}_{t} \in \mathbb{R}^{\mathit{H} \times  \mathit{W} \times 3} , \mathit{M}_{t} \in \mathbb{R}^{3 \times 4} 
\).
It is worth noting that the variation in the subscripts merely represents the change in the position and perspective of the agent when exploring the current scene continuously. 
Different subscripts may correspond to similar positions and perspectives, indicating that the agent has returned to a previously explored location. 
In embodied occupancy prediction, re-exploration of the same area should maintain global consistency and even demonstrate improved performance, akin to we humans always possessing a more comprehensive understanding of sights that have been encountered repeatedly.

We formulate the function of an embodied occupancy prediction model as follows:
\vspace{-2mm}
\begin{equation}
\label{eq2}
\begin{split}
   % \mathbf{Y}_{1} = \mathcal{F}_{embodied}(\mathit{x}_{1}), \\
   \mathbf{Y}_{t} = \mathcal{F}_{embodied}(\mathbf{Y}_{t-1}, \mathit{x}_{t}), \\
\end{split}
\end{equation}
\vspace{-4mm}

\noindent
where \(\mathcal{F}_{embodied}\) is the embodied prediction model,   \(\mathbf{Y}_{t} \in  \mathbb{R}^{\mathit{X}_{room} \times \mathit{Y}_{room} \times \mathit{Z}_{room} \times \mathit{C}}
\) refers to the current occupancy prediction of the whole scene (\(\mathbf{Y}_{0}\) is the initialization). 
\(\mathit{X}_{room}\), \(\mathit{Y}_{room}\), \(\mathit{Z}_{room}\) denote the scene dimensions. 
% which share the same coordinate system with the monocular setting.

\subsection{Local Refinement Module}
\label{sec:3.2}

Different from conventional methods that conducted feature integration in a voxelized space, 
%GaussianFormer~\cite{huang2025gaussianformer} first proposed an object-centric 3D representation to complete the 3D occupancy prediction task. 
%Motivated by this, 
we use a set of 3D semantic Gaussians to represent an indoor scene~\cite{huang2025gaussianformer}.
% and design our local and embodied occupancy prediction module based on this representation. 
In this subsection, we will first explain our local refinement module, which extracts semantic and structural features from the monocular input and integrates them to update the Gaussian-based representation of the current frustum.

\textbf{Initialization.}
We first initialize a set of semantic Gaussians to represent the current frustum.
Each semantic Gaussian \(\mathbf{G}\) is represented by a vector comprising mean \(\mathbf{m} \in \mathbb{R}^{3}\), scale \(\mathbf{s} \in \mathbb{R}^{3}\), rotation quaternion \(\mathbf{r} \in \mathbb{R}^{4}\), opacity \(\mathbf{o} \in \mathbb{R}\), and semantic logits \(\mathbf{c} \in \mathbb{R}^{\mathit{C}}\) (\(\mathit{C}\) denotes the total number of semantic categories). 
We use an embedding layer to lift each Gaussian vector \(\mathbf{G}\) to its corresponding high-dimensional feature vectors \(\mathbf{Q}\), and derive \(\mathcal{Q} =\{\mathbf{Q}_{i} \in \mathbb{R}^{m}, i=1,...,N \}\), where \(m\) is the dimension of \(\mathbf{Q}_{i}\) and \(N\) is the total number of the Gaussians.

\textbf{Depth-Aware Branch.}
Due to the variable scales and tight arrangements of indoor objects, depth ambiguity has always been one of the core challenges limiting the performance of indoor occupancy prediction models in monocular settings.
Previous work has consistently focused on how to better extract and utilize depth information from the input image. 
We design a depth-aware branch to provide more accurate and effective guidance for the update of 3D semantic Gaussians in our local refinement module.

\begin{figure}[t]
\centering
\includegraphics[width=0.40\textwidth]{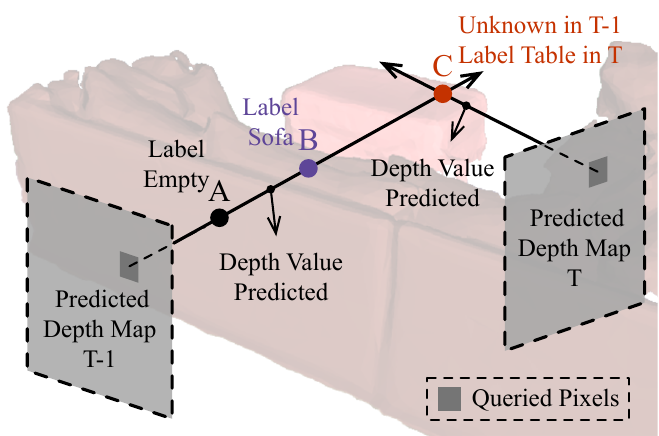}
\vspace{-3mm}
\caption{\textbf{Motivation of the depth-aware branch.}
% We use a depth-aware branch to provide local structural information for the update of each Gaussian.
Along a specific ray, Gaussians distributed in front of the true depth point are likely to model the empty semantic (A). 
Gaussians distributed behind the true depth point closely are likely to model valid semantics (B). 
Gaussians that are distributed behind the true depth point but are too far away require more information to guide their updates (C).
During the embodied exploration, the subsequent frames can make up for this lack of information in the current frame.
}
\label{fig:depthaware}
\vspace{-5mm}
\end{figure}

We first use a depth prediction network to obtain a relatively accurate depth map \(\mathit{D_{metric}}\) from input \(\mathit{I_{mono}}\). 
A naive approach can explicitly utilize this depth information when initializing the Gaussians, e.g., we can randomly sample some points from the pseudo point cloud recovered from the depth map and use these coordinates to initialize the means of some Gaussians. 
Although providing direct hints for the means of some Gaussians, this cannot exploit the potential of the depth information.
We design a simple yet effective depth-aware layer to accomplish this. 
We still uniformly initialize a number of Gaussians within the current frustum. 
For each Gaussian, we project its mean \(\mathbf{m}\) into the pixel coordinate system through the intrinsics \(\mathit{K}_{mono} \in \mathbb{R}^{3\times 3}\) and obtain the depth value \(\mathit{d}\).
The sampled depth value \(\mathit{d}\), along with the z-component \(\mathit{z}\) of the Gaussian mean in the camera coordinate system, are fed into the depth-aware layer, 
which is a multi-layer perceptron (MLP) that outputs the depth-aware feature \(\mathbf{Q}_{depth} \) for this Gaussian.
Then we add the depth-aware feature to the original feature vector \(\mathbf{Q} \), injecting additional information into the subsequent feature integration.
In this way, depth information not only affects the means of the Gaussians but also promotes the update of other properties:
\vspace{-2mm}
\begin{equation}
\label{eq3}
\begin{split}
\mathbf{Q}_{depth}=\mathcal{M}_{depthaware}((\mathit{D}_{metric}(u, v) , z),\\
\mathcal{\hat{Q}} = \{\mathbf{\hat{Q} }_{i}, i=1,...,N|\mathbf{\hat{Q}}_{i}=\mathbf{Q}_{i}+\mathbf{Q}_{i}^{depth}\},\\
\end{split}
\end{equation}
\vspace{-3mm}

\noindent
where \(\mathcal{M}_{depthaware}\) is the depth-aware layer, \((u, v)\) are pixel coordinates of each Gaussian.
We illustrate our depth-aware branch in Figure~\ref{fig:depthaware}.

\textbf{Feature Integration and Gaussian Refinement.}
Feature integration in our local refinement module includes the interactions among Gaussians as well as the interactions between image features and Gaussians.
We voxelize the Gaussian centers and conduct 3D sparse convolution on the generated grid to allow interactions among Gaussian vectors \(\mathcal{\hat{Q}} \).
We project the Gaussian centers onto the image feature map and use the deformable attention function to integrate the queried features and the Gaussian vectors \(\mathcal{\hat{Q}} \).
After the prior feature integration, these feature vectors with aggregated information will be used to obtain the update amounts \(\Delta  \mathbf{G} = (\Delta  \mathbf{m}, \Delta  \mathbf{s}, \Delta  \mathbf{r}, \Delta  \mathbf{o}, \Delta  \mathbf{c})\) of each Gaussian.
We use the update amounts \(\Delta  \mathbf{G}\) to refine the Gaussian properties:
\vspace{-2mm}
\begin{equation}
\label{eq4}
\mathbf{G}_{new} = (\Delta \mathbf{m}+\mathbf{m},  \Delta \mathbf{s}+\mathbf{s},  \Delta \mathbf{r} \otimes \mathbf{r}, \Delta \mathbf{o}+\mathbf{o}, \Delta \mathbf{c}+\mathbf{c} ),
\end{equation}
\vspace{-6mm}

\noindent
where \(\otimes\) refers to the special composition of quaternions.

We conduct the feature integration and the refinement of Gaussians multiple times.
After the final refinement, we use a Gaussian-to-voxel splatting module~\cite{huang2025gaussianformer} to obtain the final occupancy within the frustum.

\subsection{Gaussian Memory Updated Online}
%Suppose we are in an unfamiliar room,  we will first wander through it to explore the surroundings.
To explore unknown scenes, we humans continuously update the objects within the scene and their relationships to gradually construct a global scene memory.
%During this process, the objects within the scene and their relationships are continuously updated in our minds, indicating the formation of a memory regarding this scene. 
%Upon returning to the scene next time or revisiting it for further exploration
When revisiting this scene for further exploration, we use the visual information to refine this memory. 
%Indeed, the embodied occupancy prediction framework we propose in this paper operates similarly. 
Inspired by this, we design an online framework (shown in Figure~\ref{fig:framework}) and maintain a Gaussian memory for global understanding.
%In this subsection, we will elaborate on how we maintain and update the Gaussian memory used in the final embodied occupancy prediction framework.

\textbf{Memory Initialization.}
Our local refinement module initializes and updates Gaussians in the camera coordinate system, as the extrinsics in indoor scenarios are constantly changing, which will pose additional difficulties for our local module.
But in the final embodied framework, we initialize the entire scene with uniform Gaussians in the world coordinate system. 
For a novel scene to be explored, we have:
\(
\mathcal{G}_{room}=\{(\mathbf{G}_{i},\gamma_{i} ), i=1,...,N|\mathbf{G}_{i}=(\mathbf{m}_{i}, \mathbf{s}_{i}, \mathbf{r}_{i},\mathbf{o}_{i},\mathbf{c}_{i}), \gamma_{i}=0,1  \},
\)
where \(N\) refers to the number of Gaussians to initialize this scene, \(\mathbf{m}_{i}\) and \(\mathbf{r}_{i}\) are the means and rotation quaternions of these Gaussians in the world coordinate system (\(\mathbf{s}_{i}\), \(\mathbf{o}_{i}\) and \(\mathbf{c}_{i}\) maintain consistency between the world and camera coordinate systems).
We introduce an additional tag \(\gamma \) for all the Gaussians in the memory. 
When initializing a novel scene, tags of these Gaussians are set to \(0\). 
Every time we put some updated Gaussians back into the memory, their tags are set to \(1\).

\textbf{Memory Update.}
At the current step \(t\), our embodied occupancy prediction framework receives a posed visual input \(\mathit{x}_{t} = (\mathit{I}_{t}, \mathit{M}_{t})\) to perform the update. 
During the current update, we use a mask from coordinate system transformation to get all Gaussians \(\mathcal{G}_{t} \) within the current frustum from the memory.
These Gaussians will interact and be refined using a tailored confidence refinement module.
Then we detach these Gaussians and put them back into the memory.

\textbf{Confidence Refinement.}
Apart from the initial update for each scene which is akin to the local refinement, subsequent exploration involves the update of Gaussians from the Gaussian memory, among which some have been well-updated by previous frames (if we can derive an acceptable local occupancy prediction from these Gaussians, we believe that they have been “well-updated”) and some still remain random.
It is unreasonable to update these Gaussians equally. 
For those Gaussians deemed well-updated, we only need to refine them slightly based on the semantic and structural features extracted from the current image, which is exactly the essence of maintaining the Gaussian memory. 
As for those random Gaussians that have never been updated, we can directly update them with a fresh perspective.

\begin{figure}[t]
\centering
\includegraphics[width=0.475\textwidth]{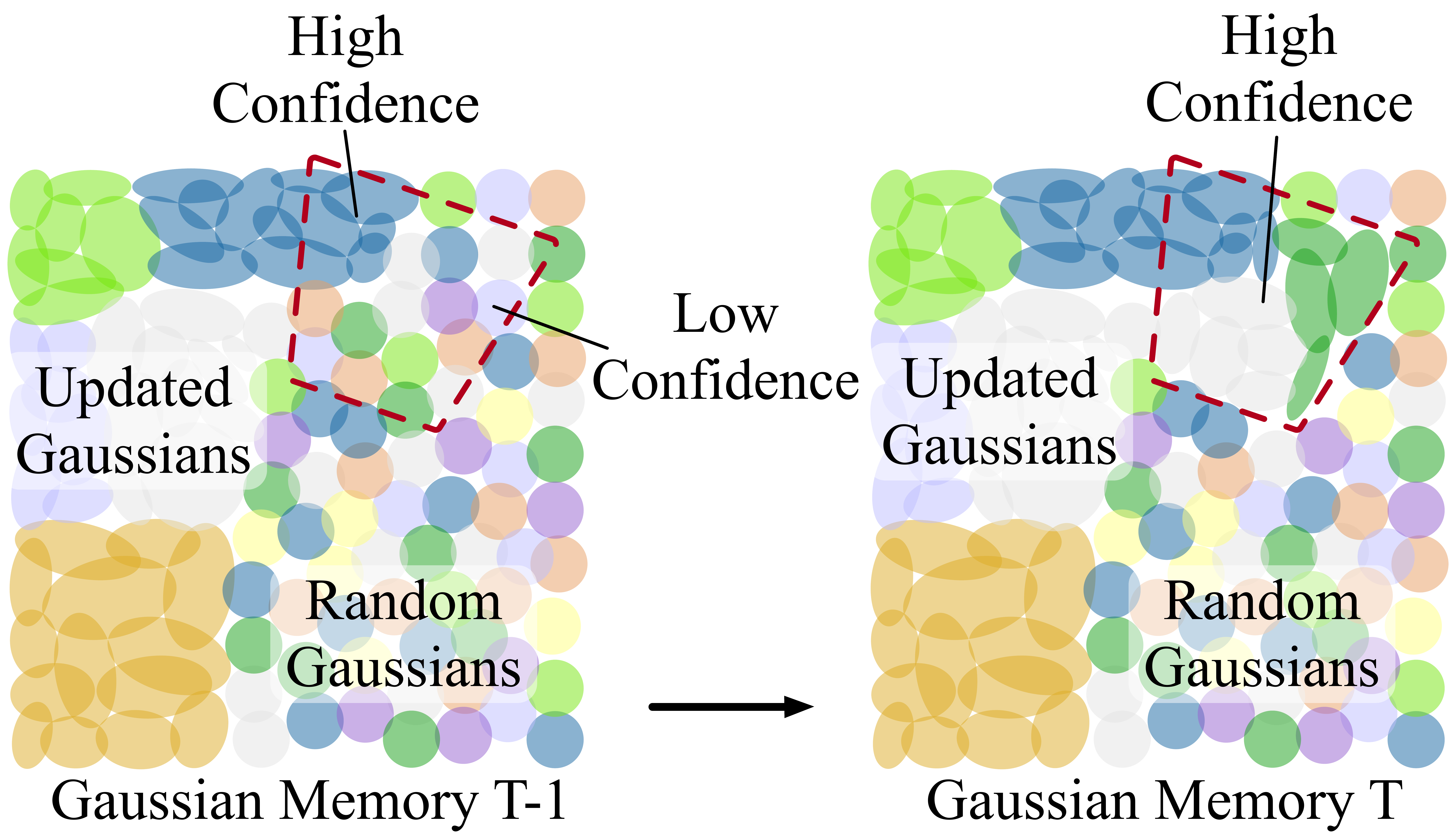}
\vspace{-7mm}
\caption{\textbf{Illustration of our Gaussian memory.}
During each update, the Gaussians within the current frustum are taken from the memory.
Confidence values of those well-updated Gaussians are used to integrate information from both the memory and the current input.
Then we put these Gaussians back into the memory.
}
\label{fig:memory}
\vspace{-5mm}
\end{figure}

To elaborate, we generate a confidence value \(\theta\) for each Gaussian taken from the memory. 
For those having been previously updated (\(\gamma=1 \)), we set their confidence values to a certain value between \(0\) and \(1\), which means we will integrate information from both the memory and the newly observed image to update these Gaussians. 
For those that have never been updated, we set their confidence values to \(0\) and follow the same refinement module in Sec.~\ref{sec:3.2}:
\vspace{-2mm}
\begin{equation}
\label{eq5}
\begin{split}
\Delta \mathbf{G}_{online} = (1-\theta ) \Delta \mathbf{G},\\
\mathbf{G}_{after}= \Delta \mathbf{G}_{online} \oplus \mathbf{G}_{before},\\
\end{split}
\end{equation}
\vspace{-4mm}

\noindent
where we use \(\oplus\) to represent the composition of rotation quaternions and the add operation of other parts.
Figure~\ref{fig:memory} illustrates how we maintain the Gaussian memory.

\textbf{Stopping Mechanism.}
We propose a simple stopping mechanism to consider a room as having been effectively explored.
At the step \(t\), we first calculate a confidence ratio \(\alpha\) to measure the exploration of the current room:
\vspace{-2mm}
\begin{equation}
\label{eq:eq6} 
\alpha = {\textstyle \sum_{i=0}^{N}} {\mathbb{I}_{{\gamma}_{i}=1}} / N, 
\end{equation}
\vspace{-5mm}

\noindent
where \(\mathbb{I}_{{\gamma}_{i}=1}\) takes the value of \(1\) if \({\gamma}_{i}=1\).
If \(\alpha\) exceeds a certain threshold we set before, the model can decide to enter an adjacent room to begin a new exploration or stay here to get a better perception of the current room.

\subsection{EmbodiedOcc: An Embodied Framework}
We present the training framework of our EmbodiedOcc model for indoor embodied occupancy prediction.
During the whole prediction process, we use the current monocular input to update our Gaussian memory in real time, which can be easily converted into 3D occupancy prediction.

We first train our local refinement module using the focal loss \(\mathit{L}_{focal}\), the lovasz-softmax loss \(\mathit{L}_{lov}\), the scene-class affinity loss \(\mathit{L}_{scal}^{geo}\) and \(\mathit{L}_{scal}^{sem}\) following RetinaNet~\cite{lin2017focal}, TPVFormer~\cite{tpvformer} and MonoScene~\cite{monoscene}. We use monocular occupancy within the frustum \(\mathbf{Y}_{mono}^{fov}\) and the corresponding ground truth \(\mathbf{Y}_{gt}^{fov}\) to compute the loss:
\vspace{-3mm}
\begin{equation}
\label{eq6}
\begin{split}
\mathcal{L}=\lambda_{1} \mathcal{L}_{focal}(\mathbf{Y}_{mono}^{fov}, \mathbf{Y}_{gt}^{fov}) 
+ \mathcal{L}_{lov}(\mathbf{Y}_{mono}^{fov}, \mathbf{Y}_{gt}^{fov}) \\
+ \mathcal{L}_{scal}^{geo}(\mathbf{Y}_{mono}^{fov}, \mathbf{Y}_{gt}^{fov})
+\mathcal{L}_{scal}^{sem}(\mathbf{Y}_{mono}^{fov}, \mathbf{Y}_{gt}^{fov}),\\
\end{split}
\end{equation}
\vspace{-4mm}

\noindent
where \(\lambda_{1}\) is a balance factor.

\begin{table*}[!t] \small
		\caption{
        \textbf{Local Prediction Performance on the Occ-ScanNet dataset.}
        }
        \vspace{-7mm}
		\small
		\setlength{\tabcolsep}{0.008\textwidth}
		\captionsetup{font=scriptsize}
		% \centering
            \begin{center}
%            \resizebox{1.0\linewidth}{!}{
		\begin{tabular}{l|c|c|c c c c c c c c c c c|c}
			\toprule
			Method
			& Input
			& {IoU}
			& \rotatebox{90}{\parbox{1.5cm}{\textcolor{ceiling}{$\blacksquare$} ceiling}} 
			& \rotatebox{90}{\textcolor{floor}{$\blacksquare$} floor}
			& \rotatebox{90}{\textcolor{wall}{$\blacksquare$} wall} 
			& \rotatebox{90}{\textcolor{window}{$\blacksquare$} window} 
			& \rotatebox{90}{\textcolor{chair}{$\blacksquare$} chair} 
			& \rotatebox{90}{\textcolor{bed}{$\blacksquare$} bed} 
			& \rotatebox{90}{\textcolor{sofa}{$\blacksquare$} sofa} 
			& \rotatebox{90}{\textcolor{table}{$\blacksquare$} table} 
			& \rotatebox{90}{\textcolor{tvs}{$\blacksquare$} tvs} 
			& \rotatebox{90}{\textcolor{furniture}{$\blacksquare$} furniture} 
			& \rotatebox{90}{\textcolor{objects}{$\blacksquare$} objects} 
			& mIoU\\
			\midrule
                TPVFormer~\cite{tpvformer} & $x^{\text{rgb}}$ & 33.39 & 6.96 & 32.97 & 14.41 & 9.10 & 24.01 & 41.49 & 45.44 & 28.61 & 10.66 & 35.37 & 25.31 & 24.94 \\
                GaussianFormer~\cite{huang2025gaussianformer} & $x^{\text{rgb}}$ & 40.91 & 20.70 & 42.00 & 23.40 & 17.40 & 27.0 & 44.30 & 44.80 & 32.70 & 15.30 & 36.70 & 25.00 & 29.93 \\
			MonoScene~\cite{monoscene} & $x^{\text{rgb}}$ & 41.60 & 15.17 & 44.71 & 22.41 & 12.55 & 26.11 & 27.03 & 35.91 & 28.32 & 6.57 & 32.16 & 19.84 & 24.62 \\
            ISO~\cite{yu2024monocular} & $x^{\text{rgb}}$ & 42.16 & 19.88 & 41.88 & 22.37 & 16.98 & 29.09 & 42.43 & 42.00 & 29.60 & 10.62 & 36.36 & 24.61 & 28.71 \\
            Surroundocc~\cite{surroundocc} & $x^{\text{rgb}}$ & 42.52 & 18.90 & 49.30 & 24.80 & 18.00 & 26.80 & 42.00 & 44.10 & 32.90 & 18.60 & 36.80 & 26.90 & 30.83 \\
            Ours & $x^{\text{rgb}}$ & \textbf{53.55} & \textbf{39.60} & \textbf{50.40} & \textbf{41.40} & \textbf{31.70} & \textbf{40.90} & \textbf{55.00} & \textbf{61.40} & \textbf{44.00} & \textbf{36.10} & \textbf{53.90} & \textbf{42.20} & \textbf{45.15} \\
			\bottomrule
		\end{tabular}
%		}
            \end{center}
\vspace{-7mm}
		\label{tab:mono_scannet}
 \end{table*}

\begin{table*}[!t] 	\small
		\caption{
        \textbf{Embodied Prediction Performance on the EmbodiedOcc-ScanNet dataset.}
        }
        \vspace{-7mm}
		\setlength{\tabcolsep}{0.007\textwidth}
		\captionsetup{font=scriptsize}
		% \centering
            \begin{center}
%            \resizebox{1.0\linewidth}{!}{
		\begin{tabular}{l|c|c|c c c c c c c c c c c|c}
			\toprule
			Method
			& Dataset
			& {IoU}
			& \rotatebox{90}{\parbox{1.5cm}{\textcolor{ceiling}{$\blacksquare$} ceiling}} 
			& \rotatebox{90}{\textcolor{floor}{$\blacksquare$} floor}
			& \rotatebox{90}{\textcolor{wall}{$\blacksquare$} wall} 
			& \rotatebox{90}{\textcolor{window}{$\blacksquare$} window} 
			& \rotatebox{90}{\textcolor{chair}{$\blacksquare$} chair} 
			& \rotatebox{90}{\textcolor{bed}{$\blacksquare$} bed} 
			& \rotatebox{90}{\textcolor{sofa}{$\blacksquare$} sofa} 
			& \rotatebox{90}{\textcolor{table}{$\blacksquare$} table} 
			& \rotatebox{90}{\textcolor{tvs}{$\blacksquare$} tvs} 
			& \rotatebox{90}{\textcolor{furniture}{$\blacksquare$} furniture} 
			& \rotatebox{90}{\textcolor{objects}{$\blacksquare$} objects} 
			& mIoU\\
			\midrule
                TPVFormer~\cite{tpvformer} & EmbodiedOcc & 35.88 & 1.62 & 30.54 & 12.03 & 13.22 & 35.47 & 51.39 & 49.79 & 25.63 & 3.60 & 43.15 & 16.23 & 25.70 \\
                SurroundOcc~\cite{surroundocc} & EmbodiedOcc & 37.04 & 12.70 & 31.80 & 22.50 & 22.00 & 29.90 & 44.70 & 36.50 & 24.60 & 11.50 & 34.40 & 18.20 & 26.27 \\
                GaussianFormer~\cite{huang2025gaussianformer} & EmbodiedOcc & 38.02 & 17.00 & 33.60 & 21.50 & 21.70 & 29.40 & 47.80 & 37.10 & 24.30 & 15.50 & 36.20 & 16.80 & 27.36 \\
			SplicingOcc & EmbodiedOcc & 49.01 & \textbf{31.60} & 38.80 & 35.50 & 36.30 & 47.10 & 54.50 & 57.20 & 34.40 & 32.50 & 51.20 & 29.10 & 40.74 \\
            EmbodiedOcc & EmbodiedOcc & \textbf{51.52} & 22.70 & \textbf{44.60} & \textbf{37.40} & \textbf{38.00} & \textbf{50.10} & \textbf{56.70} & \textbf{59.70} & \textbf{35.40} & \textbf{38.40} & \textbf{52.00} & \textbf{32.90} & \textbf{42.53} \\
            % SplicingOcc & EmbodiedOcc-mini & 48.75 & \textbf{29.00} & 37.60 & 37.30 & \textbf{26.80} & 44.50 & \textbf{65.90} & 52.70 & 40.80 & \textbf{36.60} & 54.50 & \textbf{27.90} & 41.24 \\
            % EmbodiedOcc & EmbodiedOcc-mini & \textbf{50.78} & 22.10 & \textbf{43.70} & \textbf{39.00} & 26.60 & \textbf{45.00} & 63.70 & \textbf{54.40} & \textbf{43.90} & 34.70 & \textbf{55.30} & 27.60 & \textbf{41.45} \\
			\bottomrule
		\end{tabular}
%		}
            \end{center}
            \vspace{-9mm}
		\label{tab:global_main}
 \end{table*}

We then use the trained local module to train our EmbodiedOcc. 
For efficient training, we initialize the Gaussian memory before the first update and compute the current loss following the equation~\ref{eq6} after each update. 
To ensure consistency, the local occupancy ground truth used here is obtained from the occupancy of the whole scene.
After a certain number of updates, we re-initialize the memory and come to the next scene. 
Trained with such a pipeline, our EmbodiedOcc can effectively perform the embodied occupancy prediction task while ensuring consistency within the same scene. 
We expect that our EmbodiedOcc can have an improving prediction with continuous exploration rather than undermining previous predictions when encountering parts that have been explored before. 
Therefore, we conduct some tailored tests to validate the capability of our model.

\section{Experiments}

\subsection{EmbodiedOcc-ScanNet Benchmark}
%We construct an EmbodiedOcc-ScanNet benchmark based on the locally annotated Occ-Scannet dataset~\cite{dai2017scannet, yu2024monocular}. 
%We explain our benchmark in detail in three parts: task descriptions, datasets, and evaluation metrics we use.

\textbf{Task Descriptions.}
We conducted two tasks to evaluate our EmbodiedOcc framework: local occupancy prediction and embodied occupancy prediction. 
Local occupancy prediction shares the same setting with previous works, which accept monocular image as input and obtain the occupancy within the current frustum.
Embodied occupancy prediction accepts real-time visual inputs continuously and updates the occupancy of the current scene online.
The visual input at a certain step \(t\) during embodied occupancy prediction is still monocular, which is a more challenging setting compared with multi-view input or input with 3D information.

\textbf{Datasets.}
We trained and evaluated our local refinement module on the Occ-ScanNet dataset~\cite{yu2024monocular}, which provides frames in \(60\times60\times36\) voxel grids (a \(4.8m\times4.8m\times2.88m\) box in front of the camera). 
% In the local occupancy prediction task, we used the Occ-ScanNet dataset~\cite{yu2024monocular} which provides frames in \(60\times60\times36\) voxel grids (a \(4.8m\times4.8m\times2.88m\) box in front of the camera). 
These frames are labeled with 12 semantics, including 11 for valid semantics (ceiling, floor, wall, window, chair, bed, sofa, table, tvs, furniture, objects) and 1 for empty space. 
% We trained and evaluated our local refinement module on this dataset.

Based on this dataset, we reorganized an EmbodiedOcc-ScanNet dataset to train and evaluate our EmbodiedOcc framework~\cite{silberman2012indoor, yu2024monocular}. 
% During the training and evaluation of our EmbodiedOcc framework, we have to ensure that scenes in the training set are different from those in the evaluating set. 
% So we split the scenes again and obtained our final EmbodiedOcc-ScanNet dataset, which comprises 537/137 scenes in the train/val splits.
Our EmbodiedOcc-ScanNet comprises 537/137 scenes in the train/val splits.
Each scene consists of 30 posed frames with their corresponding local occupancies.
The resolutions of global occupancy of each scene are calculated by \(\mathit{l}_{x}\times\mathit{l}_{y}\times\mathit{l}_{z}/(0.08m)^3\), where \(\mathit{l}_{x}\times\mathit{l}_{y}\times\mathit{l}_{z}\) is the range of this scene in the world coordinate system.
In addition, we associate grid points that can be projected onto the camera plane for each frame as the global mask of this frame.
By splicing the global mask of all processed frames, we can easily obtain the occupancy ground truth of the explored part in the current scene.

Apart from Occ-ScanNet and EmbodiedOcc-ScanNet datasets in the original scale, we sampled a small set from the EmbodiedOcc-ScanNet dataset as the EmbodiedOcc-ScanNet-mini dataset which comprises 64/16 scenes in the train/val splits. 
We sampled from the Occ-ScanNet dataset accordingly and obtained an Occ-ScanNet-mini2 dataset, which comprises 5504/2376 frames in the train/val splits.
We conducted the local occupancy prediction task on the Occ-ScanNet and Occ-ScanNet-mini2 datasets and conducted the embodied prediction task on the EmbodiedOcc-ScanNet and EmbodiedOcc-ScanNet-mini datasets.

\textbf{Evaluation Metrics.}
We use mIoU and IoU as the evaluation metrics.
For local occupancy prediction, we calculate the mIoU and IoU using the occupancy within the current frustum (same with the evaluation in ISO~\cite{yu2024monocular}).
For embodied occupancy prediction, we calculate the mIoU and IoU using the global occupancy of the current scene.
It is worth mentioning that the global occupancy used here is the union of the frustums corresponding to 30 frames of each scene, which represents the region that has been explored.

\subsection{Implementation Details}
\textbf{Local Refinement Module.}
Following existing works~\cite{huang2025gaussianformer, yu2024monocular}, we use a pre-trained EfficientNet-B7~\cite{tan2019efficientnet} to initialize the image encoder in our local module.
The depth prediction network used in the depth-aware branch is a fine-tuned DepthAnything-V2 model~\cite{yang2024depth} that remains frozen during the training, and the depth-aware layer is a 3-layer MLP.
% We use 3 hybrid blocks to implement feature integration and gaussian refinement. 
% The other parts of our local occupancy prediction module follow the GaussianFormer\cite{huang2025gaussianformer}.
The resolutions of the monocular input are set to \(480\times640\) and the number of Gaussians used to conduct the local prediction is 16200.
We utilize the AdamW~\cite{loshchilov2017adamw} optimizer with a weight decay of 0.01.
The learning rate warms up in the first 1000 iterations to a maximum value of 2e-4 and decreases according to a cosine schedule~\cite{loshchilov2016cosineanneal}.
We train our local refinement module for 10 epochs using 8 NVIDIA GeForce RTX 4090 GPUs on the Occ-ScanNet dataset and 20 epochs on the Occ-ScanNet-mini2 dataset.

\textbf{EmbodiedOcc Framework.}
We initialize the Gaussians with a 0.16 m interval to represent a novel scene.
For each update, the confidence value \(\theta\) of well-updated Gaussians is set to 0 in the first two refinement layers (frozen) and 0.5 in the final refinement layer.
We train our EmbodiedOcc for 5 epochs using 8 NVIDIA GeForce RTX 4090 GPUs on the EmbodiedOcc-ScanNet dataset and 20 epochs using 4 NVIDIA GeForce RTX 4090 GPUs on the EmbodiedOcc-ScanNet-mini dataset.
The maximum value of the learning rate is set to 2e-4 using 8 GPUs and 1e-4 using 4 GPUs.
The other settings remain the same with the training of the local refinement module.

\begin{table} \tablesize
		\caption{
        \textbf{Look-Back Prediction \textit{vs} First-Time Prediction.}
        For \(\mathrm{K}=k\), we simply select \(0,1,...,\mathrm{k-1}\)th frames to evaluate our EmbodiedOcc framework and the occupancy ground truth used here is the union of the \(k\) frustums. \(\mathrm{K}\) was set to 3/5/8.
        }
        \vspace{-3mm}
		\small
\setlength{\tabcolsep}{8pt}
%		\captionsetup{font=scriptsize}
		\centering
%            \resizebox{1.0\linewidth}{!}{
		\begin{tabular}{l|c|c|c c}
			\toprule
            {Mode} & {K} & {Frame List} & IoU & mIoU \\
			\midrule
			First-Time & 3 & \([0,1,2]\) & 49.39 & 39.32 \\
                Look-Back & 3 & \([0,1,2,1,0]\) & \textbf{49.52} & \textbf{40.09} \\
                First-Time & 5 & \([0,...,4]\) & 50.13 & 40.03 \\
                Look-Back & 5 & \([0,...,3,4,3,...,0]\) & \textbf{50.64} & \textbf{40.98} \\
                First-Time & 8 & \([0,...,7]\) & 50.94 & 40.86 \\
                Look-Back & 8 & \([0,...,6,7,6,...,0]\) & \textbf{51.14} & \textbf{41.17} \\
			\bottomrule
		\end{tabular}
%		}
\vspace{-4mm}
		\label{tab:lookback}
 \end{table}

\begin{figure}
    \centering
    \begin{minipage}{0.23\textwidth}
        \centering
        \includegraphics[width=\linewidth]{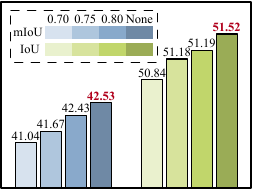}
        \vspace{-6mm}
        \caption{\textbf{Performance with different stopping ratios.}} 
        \label{fig:tab_stop_ratios}
    \end{minipage}
    \hfill
    \begin{minipage}{0.23\textwidth}
        \centering
        \includegraphics[width=\linewidth]{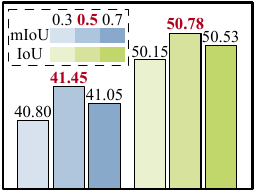}
        \vspace{-6mm}
        \caption{\textbf{Ablation study of the confidence refinement.}}
        \label{fig:tab_confidence}
    \end{minipage}
     \vspace{-6mm}
\end{figure}

% \begin{table} \tablesize
% %\vspace{-4mm}
% \caption{\textbf{Performance with different stopping ratios.} 
% }
% \vspace{-3mm}
% \small
% \centering
% %\resizebox{1.0\linewidth}{!}{
% \setlength{\tabcolsep}{12pt}
% \begin{tabular}{c|c|c|c}
% \toprule
% Threshold & IoU & mIoU & \#Seen Frames \\
% \midrule
% 0.7 & 50.84 & 41.04 & 22.0 \\
% 0.75 & 51.18 & 41.67 & 24.2 \\
% 0.8 & 51.19 & 42.43 & 26.6 \\
% None & 51.52 & 42.53 & 30.0 \\
% \bottomrule
% \end{tabular}
% %}
% \vspace{-4mm}
% \label{tab:stop_ratio}
% \end{table}

\subsection{Main Results}

\textbf{Local Occupancy Prediction.}
We evaluated our local refinement module on the Occ-ScanNet dataset~\cite{yu2024monocular}.
As shown in Table~\ref{tab:mono_scannet}, the results indicate that our local refinement module outperforms ISO~\cite{yu2024monocular}.
We also implemented several state-of-the-art driving scene methods~\cite{tpvformer, surroundocc, huang2025gaussianformer} on this benchmark and our local refinement module outperforms them by a large margin.
This is because they mainly focus on the coarse layout (e.g., positions of objects) while indoor scenes require modeling of the fine-grained structure (e.g., shapes of objects).

\textbf{Embodied Occupancy Prediction.}
We assessed the occupancy prediction for the entire scene after processing 30 frames, and the ground truth for calculating IoU and mIoU is the union of the frustums. 
We spliced the local occupancy obtained from our local module to serve as the main baseline (referred to as SplicingOcc), as our local module has achieved the best local performance to date. 
It can be observed in Table~\ref{tab:global_main} that our EmbodiedOcc exhibits superior prediction of the scene, which is achieved through the integration of different views.
We also compared our EmbodiedOcc with the driving scene methods mentioned before (we obtained their embodied results by voting from different local predictions).
Their poor results are due to ignoring the continuity of the observations without a global memory.

\subsection{Experimental Analysis}

\textbf{Effect of Continuous Online Updating.}
We expect EmbodiedOcc to have better performance when encountering parts that have been explored before, and thus, we designed a Look-Back evaluation on the EmbodiedOcc-ScanNet dataset. 
Specifically, after processing \(\mathrm{K}\) frames, we direct the model to reprocess the last \(\mathrm{K-1}\) frames.
By comparing this Look-Back result with the First-Time prediction, we verified that our EmbodiedOcc has met our expectations as shown in Table~\ref{tab:lookback}.

\begin{table} \tablesize
\caption{\textbf{Analysis of the model design.} 
}
\vspace{-3mm}
\centering
\resizebox{1.0\linewidth}{!}{
\setlength{\tabcolsep}{2pt}
\begin{tabular}{c|ccc|cc|cc}
\toprule
\multirow{2}{*}{Method} & \multirow{2}{*}{Gaussian} &\multirow{2}{*}{Structure} & \multirow{2}{*}{Memory} & \multicolumn{2}{c|}{Local Prediction} & \multicolumn{2}{c}{Embodied Prediction} \\
 & & & & IoU & mIoU & IoU & mIoU \\
\midrule
EmbodiedOcc-Voxel & $\times$ & $\checkmark$ & $\checkmark$ & 47.50 & 38.12 & 37.53 & 26.99 \\
EmbodiedOcc w/o memory & $\checkmark$ & $\checkmark$ & $\times$ & \textbf{53.55} & \textbf{45.15} & 49.01 & 40.74 \\
EmbodiedOcc & $\checkmark$ & $\checkmark$ & $\checkmark$ & \textbf{53.55} & \textbf{45.15} & \textbf{51.52} & \textbf{42.53} \\
\bottomrule
\end{tabular}
}
\vspace{-4mm}
\label{tab:model_design}
\end{table}

\begin{table}[t]
\caption{\textbf{Analysis of the depth-aware branch.} 
}
\vspace{-3mm}
\small
\centering
\resizebox{1.0\linewidth}{!}{
\setlength{\tabcolsep}{2pt}
\begin{tabular}{c|c|cc|cc}
\toprule
 \multirow{2}{*}{Branch Type} & \multirow{2}{*}{Depth Estimation Module} &  
\multicolumn{2}{c|}{Local Prediction} & 
\multicolumn{2}{c}{Embodied Prediction} \\
 & & IoU & mIoU & IoU & mIoU \\
\midrule
Depth-aware branch & DepthAnything-V2 & \textbf{53.93} & \textbf{46.20} & \textbf{50.78} & \textbf{41.45} \\
No-depth branch & / & 48.15 & 40.07 & 37.52 & 30.73 \\
Naive-depth branch & DepthAnything-V2 & 50.32 & 42.73 & / & / \\
Depth-aware branch & IndoorDepth & 51.24 & 43.87 & 46.42 & 37.78 \\
% Depth-aware branch & 8100 & 0.01 & 0.08 & (0.20, 0.20, 0.20) & 50.47 & 42.82 & 46.24 & 37.99 \\
% Depth-aware branch & 16200 & 0.01 & 0.20 & (0.16, 0.16, 0.16) & 51.57 & 43.74 & 48.09 & 38.40 \\
\bottomrule
\end{tabular}}
\vspace{-4mm}
\label{tab:ablation2}
\end{table}

\begin{table}
\caption{\textbf{Analysis of the Gaussian parameters.} 
}
\vspace{-3mm}
\small
\centering
\resizebox{1.0\linewidth}{!}{
\setlength{\tabcolsep}{2pt}
\begin{tabular}{c|cc|c|cc|cc}
\toprule
Gaussian Number & 
 \multicolumn{2}{c|}{Gaussian Scale} & Gaussian Interval(m) & 
\multicolumn{2}{c|}{Local Prediction} & 
\multicolumn{2}{c}{Embodied Prediction} \\
(In local box) & Min(m) & Max(m) & (In global scene) & IoU & mIoU & IoU & mIoU \\
\midrule
16200 & 0.01 & 0.08 & (0.16, 0.16, 0.16) & \textbf{53.93} & \textbf{46.20} & \textbf{50.78} & \textbf{41.45} \\
8100 & 0.01 & 0.08 & (0.20, 0.20, 0.20) & 50.47 & 42.82 & 46.24 & 37.99 \\
16200 & 0.01 & 0.20 & (0.16, 0.16, 0.16) & 51.57 & 43.74 & 48.09 & 38.40 \\
\bottomrule
\end{tabular}}
\vspace{-4mm}
\label{tab:ablation3}
\end{table}

% \begin{table}[t] \tablesize
% \setlength{\tabcolsep}{0.016\linewidth}
% \caption{\textbf{Ablation study of the confidence refinement.} 
% }
% \vspace{-3mm}
% \small
% \centering
% %\resizebox{1.0\linewidth}{!}{
% \setlength{\tabcolsep}{6pt}
% \begin{tabular}{c|c|c|cc}
% \toprule
%  Frozen & Confidence & 
%  \multirow{2}{*}{Coefficient \(\theta\)} & 
% \multicolumn{2}{c}{Embodied Prediction} \\
% Layers & Layers &  & IoU & mIoU \\
% \midrule
% 2 & 1 & 0.5 & \textbf{50.78} & \textbf{41.45} \\
% 3 & 0 & 0.5 & 48.33 & 39.44 \\
% 1 & 2 & 0.5 & 50.36 & 40.99 \\
% 0 & 3 & 0.5 & 50.18 & 40.28 \\
% 2 & 1 & 0.7 & 50.53 & 41.05 \\
% 2 & 1 & 0.3 & 50.15 & 40.80 \\
% \bottomrule
% \end{tabular}
% %}
% \vspace{-7mm}
% \label{tab:ablation4}
% \end{table}

\begin{table} \tablesize
\caption{
    \textbf{Runtime decomposition.}}
    \vspace{-3mm}
   \small
    \centering
%    \resizebox{1.0\linewidth}{!}{
\setlength{\tabcolsep}{1pt}
    \begin{tabular}{c|cc|cc}
    \toprule
    Scene level (ms) &  Scene init.  & {6.626}  &  Occ head  & 39.635                              \\ 
    \midrule
    \multirow{3}{*}{\begin{tabular}[c]{c}Frame level (ms) \end{tabular}} & Load memory   & \multicolumn{1}{c|}{0.973} & Depth aware   & 1.816 \\
& Img backbone  & \multicolumn{1}{c|}{61.478} & GS Encoder    & 14.761 \\
& Depthanything & \multicolumn{1}{c|}{34.687} & Update memory & 0.474 \\ 
    \bottomrule         
    \end{tabular}
%    }
    \vspace{-6mm}
    \label{tab:runtime}
\end{table}

\begin{figure*}[t]
\centering
\includegraphics[width=\textwidth]{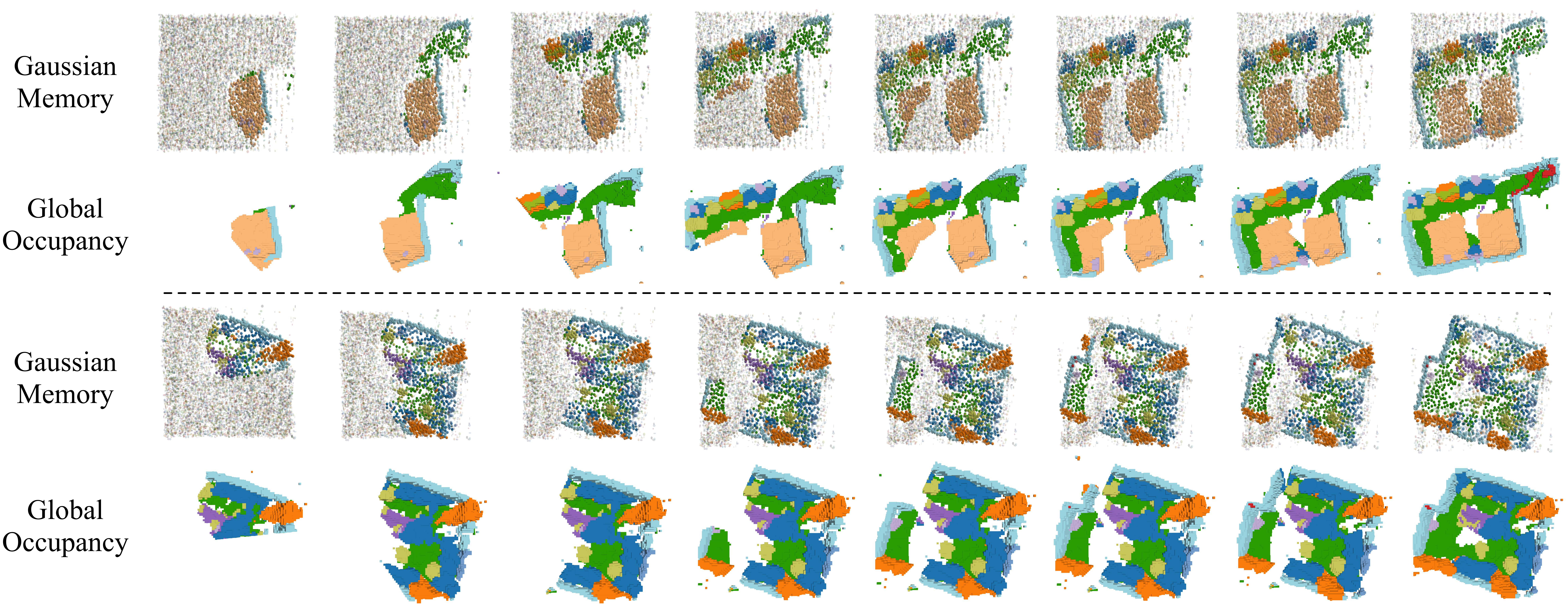}
\vspace{-8mm}
\caption{\textbf{Visualization of the embodied occupancy prediction.}
We visualize the update of Gaussian memory and corresponding global occupancy.
As the Gaussians transition from random to ordered, the occupancy of the current scene becomes more accurate and complete.
}
\label{fig:global}
\vspace{-6mm}
\end{figure*}

\begin{figure}[t]
\centering
\includegraphics[width=0.45\textwidth]{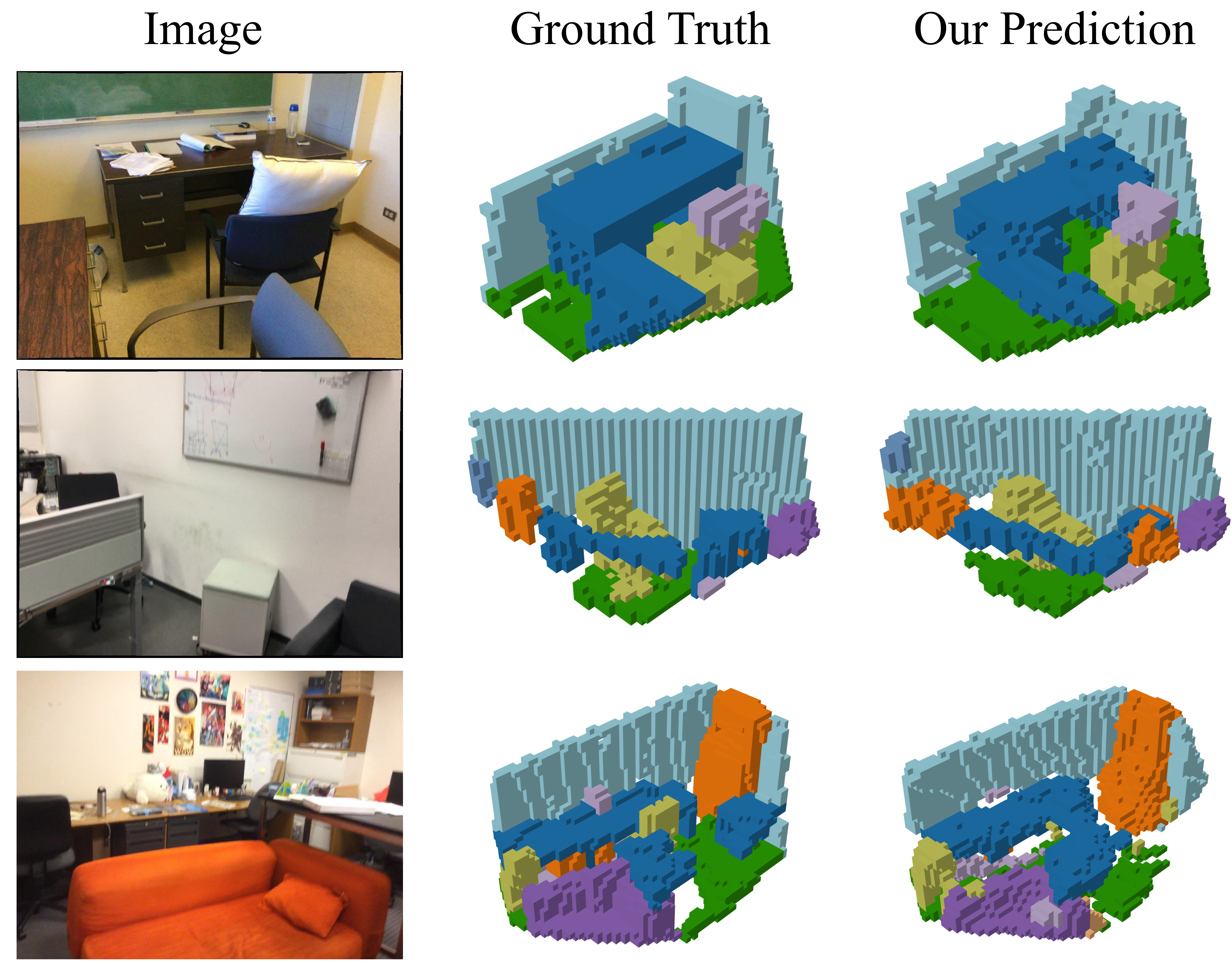}
\vspace{-3mm}
\caption{\textbf{Visualization of local occupancy prediction.}
% The input image is displayed on the left and our prediction on the right, while the ground truth is shown in the middle.
}
\label{fig:mono}
\vspace{-8mm}
\end{figure}

\textbf{Effect of the Stopping Mechanism.}
We use Figure~\ref{fig:tab_stop_ratios} to show the effectiveness of our stopping mechanism. 
% We use Table~\ref{tab:stop_ratio} to show the effectiveness of our stopping mechanism. 
The ground truth used here for calculating IoU and mIoU is the union occupancy of the 30 frustums in a global scene.
We observed that using a larger threshold results in more observations and better performance.

\textbf{Analysis of the Confidence Refinement.}
During each update, local Gaussians are refined through three refinement layers. 
For Gaussians that have been updated before, we froze the first two layers and updated them in the last refinement layer when training our EmbodiedOcc. 
Figure~\ref{fig:tab_confidence} on the Occ-ScanNet-mini2 and the EmbodiedOcc-ScanNet-mini datasets shows the impact of different confidence values (determines the coefficient of each \(\Delta \mathbf{G}\)).
% Table~\ref{tab:ablation4} on the Occ-ScanNet-mini2 and the EmbodiedOcc-ScanNet-mini datasets shows the impact of varying numbers of frozen refinement layers and different confidence values (determines the coefficient of each \(\Delta \mathbf{G}\)).
%on the embodied occupancy prediction task. 
We observe that moderate updates to those previously processed Gaussians yield the best embodied occupancy prediction.

\textbf{Analysis of the Model Design.}
The essence of our EmbodiedOcc is the explicit Gaussian memory.
We adopt object-centric Gaussians instead of grid-based voxels since Gaussians are more flexible for local-global interaction.
We implemented a voxel version of our EmbodiedOcc and evaluated it on our benchmark.
As shown in Table~\ref{tab:model_design}, the satisfactory local yet poor embodied performance of EmbodiedOcc in the voxel version verified our conclusion. 
Results in Table~\ref{tab:model_design} were evaluated on the Occ-ScanNet and EmbodiedOcc-ScanNet datasets.

\textbf{Analysis of the Depth-Aware Branch.}
We analyze the effect of our depth-aware branch in Table~\ref{tab:ablation2} using the Occ-ScanNet-mini2 and the EmbodiedOcc-ScanNet-mini datasets.
We find that depth information will significantly benefit the local and embodied occupancy prediction.
As shown in the second row, without the assistance of depth information, the performance of embodied occupancy prediction drops sharply. 
This indicates that the update of Gaussians within the current frustum may corrupt previous predictions without the guidance of depth information.
The results in the third row suggest that the depth-aware branch we employ is more reasonable compared to the naive method of directly initializing a portion of Gaussians with the pseudo point cloud recovered from the predicted depth map, the latter also poses difficulties for the initialization of global Gaussians so we do not provide the embodied results.
Besides, we replaced DepthAnything-V2 with IndoorDepth~\cite{fan2023deeper} in the last row to prove that our depth-aware branch does not rely on a specific depth prediction network.

\textbf{Analysis of the Gaussian Parameters.}
We analyze the effect of different Gaussian parameters in Table~\ref{tab:ablation3} using the Occ-ScanNet-mini2 and the EmbodiedOcc-ScanNet-mini datasets.
We see that decreasing the number or increasing the scale of the Gaussians can lead to a decrease in performance during local and embodied occupancy prediction.
This is closely related to the physical properties of Gaussians. 
Gaussians initialized too sparse may lead to holes in occupancy prediction, while Gaussians with too large scale will overlap and influence each other which is also detrimental to the correct prediction of occupancy.

\textbf{Runtime Analysis.}
%on scene 0687-00
We present in Table~\ref{tab:runtime} a runtime analysis on scene 0687-00 from the EmbodiedOcc-ScanNet dataset.
The runtime decomposition details show that our method is efficient while the main bottleneck is the image and depth backbones, suggesting that the overall runtime of our EmbodiedOcc can be further reduced.

\textbf{Visualizations.}
%We conducted qualitative visualization to demonstrate the performance of our EmbodiedOcc in Figure~\ref{fig:global}.
%We also conducted visualization to demonstrate the performance of our local occupancy prediction module in Figure~\ref{fig:mono}.
Figure~\ref{fig:global} and \ref{fig:mono} visualize the global and local predictions, respectively.
Our model demonstrates reasonable local perception ability and further achieves good online prediction with the Gaussian memory.
Due to space limitations, we will use a more diverse set of samples to further show the visual effect of our EmbodiedOcc in the supplementary material.

\section{Conclusion}

In this paper, we have presented an embodied 3D occupancy prediction task and proposed a Gaussian-based EmbodiedOcc framework accordingly.
Our EmbodiedOcc maintains an explicit Gaussian memory of the current scene and updates this memory during the exploration of this scene.
Both quantitative and visualization results have shown that our EmbodiedOcc outperforms existing methods in terms of local occupancy prediction and accomplishes the embodied occupancy prediction task with high accuracy and strong expandability.
We believe that our EmbodiedOcc paves the way for enabling active agents to conduct accurate and flexible embodied occupancy prediction.

\section*{Acknowledgements}

We would like to thank Tianyu Hu for her valuable assistance with the experiments.
This work was supported in part by the National Natural Science Foundation of China under Grant 62125603, Grant 62321005, and Grant 62336004, and in part by the Beijing Natural Science Foundation under Grant No. L247009.

% \appendix
% \input{sec/X_suppl}

{
    \small
    \bibliographystyle{ieeenat_fullname}
    \bibliography{main}
}

\end{document}